\documentclass[10pt,twocolumn,letterpaper]{article}

\usepackage[pagenumbers]{cvpr} 

\usepackage{url}
\usepackage{graphicx}
\usepackage{amsmath}
\usepackage{amssymb}
\usepackage{booktabs}
\newcommand{\m}[1]{\mathcal{#1}}
\usepackage{xcolor}
\usepackage[accsupp]{axessibility} 
\usepackage{enumitem}

\makeatletter
\usepackage[pagebackref,breaklinks,colorlinks]{hyperref}

\usepackage[capitalize]{cleveref}
\crefname{section}{Sec.}{Secs.}
\Crefname{section}{Section}{Sections}
\Crefname{table}{Table}{Tables}
\crefname{table}{Tab.}{Tabs.}

\def\cvprPaperID{2170} 
\def\confName{CVPR}
\def\confYear{2023}

\begin{document}


\title{LightPainter: Interactive Portrait Relighting with Freehand Scribble}

\author{%
  Yiqun~Mei$^{1}$\quad He Zhang$^{2}$\quad Xuaner Zhang$^{2}$\quad Jianming Zhang$^{2}$\quad Zhixin Shu$^{2}$\quad Yilin Wang$^{2}$\quad\\ Zijun Wei$^{2}$\quad Shi Yan$^{2}$\quad HyunJoon Jung$^{2}$\quad Vishal M.~Patel$^{1}$ \\ \\
  {\small$^{1}$Johns Hopkins University\quad \quad $^{2}$Adobe Inc.}
}

\twocolumn[{%
\renewcommand\twocolumn[1][]{#1}%
\maketitle
\vspace{-9mm}
\begin{center}
    \centering
   
    \includegraphics[ width=\textwidth]{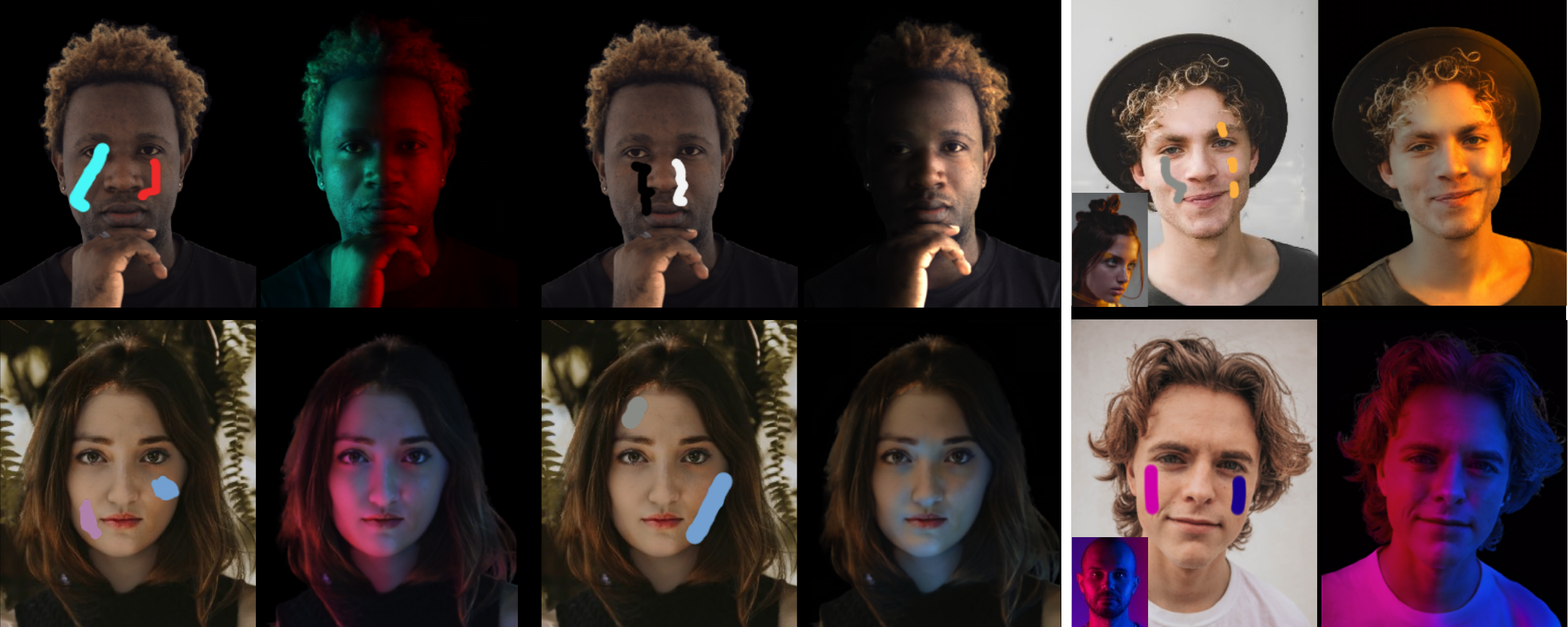}
    \vspace{-6mm}
    \captionof{figure}{LightPainter is an interactive lighting editing system that takes in an input image with freehand scribbles drawn on top and renders the correspondingly relit portrait. 
    It enables creative portrait lighting editing (left) and allows users to reproduce a target lighting effect with ease (right).
    }
   \label{fig:teaser} 
\end{center}%
}]
\begin{abstract}

Recent portrait relighting methods have achieved realistic results of portrait lighting effects given a desired lighting representation such as an environment map. However, these methods are not intuitive for user interaction and lack precise lighting control. We introduce LightPainter, a scribble-based relighting system that allows users to interactively manipulate portrait lighting effect with ease. This is achieved by two conditional neural networks, a delighting module that recovers geometry and albedo optionally conditioned on skin tone, and a scribble-based module for relighting. To train the relighting module, we propose a novel scribble simulation procedure to mimic real user scribbles, which allows our pipeline to be trained without any human annotations. We demonstrate high-quality and flexible portrait lighting editing capability with both quantitative and qualitative experiments. User study comparisons with commercial lighting editing tools also demonstrate consistent user preference for our method. 
\end{abstract}

\section{Introduction}

Lighting is a fundamental aspect of portrait photograph, as lights shape the reality, and give the work depth, colorfulness and excitement. Professional photographers~\cite{schriever1909complete,grey2014master} spend hours designing lighting such that shadow and highlight are distributed accurately on the subject to achieve the desired photographic look. Getting the exact lighting setups requires years of training, expensive equipment, environment setup, timing, and costly teamwork.
Recently, portrait relighting techniques~\cite{pandey2021total, shu2017portrait, sun2019single, zhou2019deep, hou2022face,zhang2020portrait,wang2020single,yeh2022learning,Hou_2021_CVPR} allow users to apply a different lighting condition to a portrait photo. These methods require a given lighting condition: some use an exemplar image \cite{shu2017portrait, shih2014style}, which lacks precise lighting control and requires exhaustive image search to find the specific style; some use a high dynamic range (HDR) environment maps~\cite{wang2020single,sun2019single,pandey2021total,yeh2022learning} that is difficult and unintuitive to interpret or edit.

Hand-drawn sketches and scribbles have been shown to be good for user interaction and thus are widely used in various image editing applications\cite{chen2018sketchygan,zeng2022sketchedit,nazeri2019edgeconnect,elder1998image,dekel2018sparse,olszewski2020intuitive}.
Inspired by this, we propose \textit{\textbf{LightPainter}}, a scribble-based interactive portrait relighting system. As shown in Figure ~\ref{fig:teaser}, LightPainter is an intuitive and flexible lighting editing system that only requires casual scribbles drawn on the input. 
Unlike widely-used lighting representations such as environment maps and spherical harmonics, it is non-trivial to interpret free-hand scribbles as lighting effects for a number of challenges. 

The first challenge is simulating scribbles to mimic real free-hand input as it is impractical to collect a large number of human inputs. In addition, unlike other sketch-based editing tasks~\cite{chen2018sketchygan,zeng2022sketchedit,nazeri2019edgeconnect, elder1998image,dekel2018sparse,olszewski2020intuitive} where sketches can be computed from edges or orientation maps, there is no conventional way to connect scribbles with lighting effects. To address such challenge, we propose a scribble simulation algorithm that can generate a diverse set of synthetic scribbles that mimic real human inputs. For an interactive relighting task, scribbles should be flexible and expressive: easy to draw and accurately reflecting the lighting effect, such as changes in local shading and color. Compared to a shading map, scribbles are often ``incomplete'': users tend to sparsely place the scribbles on a few key areas on the face. Therefore, we propose to use a set of locally connected ``shading stripes'' to describe local shading patterns, including shape, intensity, and color, and use them to simulate scribbles. To this end, we simulate scribbles by starting from a full shading map and applying a series of operations to generate coarse and sparse shading stripes. We show that training with our synthetic scribbles enables the system to generalize well to real user scribbles from human inputs, with which our model can generate high-quality results with desirable lighting effects.


The second challenge is how to effectively use local and noisy scribbles to robustly represent portrait lighting that is often a global effect. LightPainter uses a carefully designed network architecture and training strategy to handle these discrepancies. Specifically, we introduce a two-step relighting pipeline to process sparse scribbles. The first stage produces a plausible completion of the shading map from the input scribbles and the geometry; the second stage refines the shading and renders the appearance with a learned albedo map. We propose a carefully designed neural network with an augmented receptive field. Compared with commonly-used UNet for portrait relighting~\cite{ pandey2021total,wang2020single,Hou_2022_CVPR,nestmeyer2020learning}, our design can better handle the sparse scribbles and achieve geometry-consistent relighting.


Last, there is one major challenge in portrait relighting that originates from the ill-posed nature of the intrinsic decomposition problem. That is to decouple albedo and shading from an image. It is also difficult to address with a learning framework due to the extreme scarcity of realistic labeled data and infinite possible lighting conditions for a scene. In the context of portrait relighting, it means recovering the true skin tone of a portrait subject is very challenging~\cite{weir2022deep,Feng:TRUST:ECCV2022}. Instead of trying to collect a balanced large-scale light-stage~\cite{debevec2000acquiring} dataset to capture the continuous and subtle variations in different skin tones, we propose an alternative solution dubbed \textit{SkinFill}. We draw inspiration from the standard makeup routine and design SkinFill to allow users to specify skin tone in our relighting pipeline. We use a \textit{tone map}, a per-pixel skin tone representation, to condition the albedo prediction to follow the exact skin tone as desired. This also naturally enables additional user control at inference time. 

Similar to prior work~\cite{pandey2021total,zhang2021neural,sun2019single}, we train our system with a light stage~\cite{debevec2000acquiring} dataset. With our novel designs, LightPainter is a user-friendly system that enables creative and interactive portrait lighting editing. We demonstrate the simple and intuitive workflow of LightPainter through a thorough user study. We show it generates relit portraits with superior photo-realism and higher fidelity compared to state-of-the-art methods.  We summarize our contributions as follows:

\begin{itemize}
    \item We propose LightPainter, a novel scribble-based portrait relighting system that offers flexible user control, allowing users to easily design portrait lighting effects.
    \item We introduce a novel scribble simulation algorithm that can automatically generate realistic scribbles for training. Combining it with a carefully designed neural relighting module, our system can robustly generalize to real user input.
    \item We introduce SkinFill to allow users to specify skin tone in the relighting pipeline, which allows data-efficient training and offers additional control to address potential skin tone data bias.

\end{itemize}

%

\section{Related Work}
\begin{figure*}[t!]
	\centering
	\includegraphics[clip, trim=0.1cm 6.3cm 0.1cm 0cm, width=0.97\textwidth]{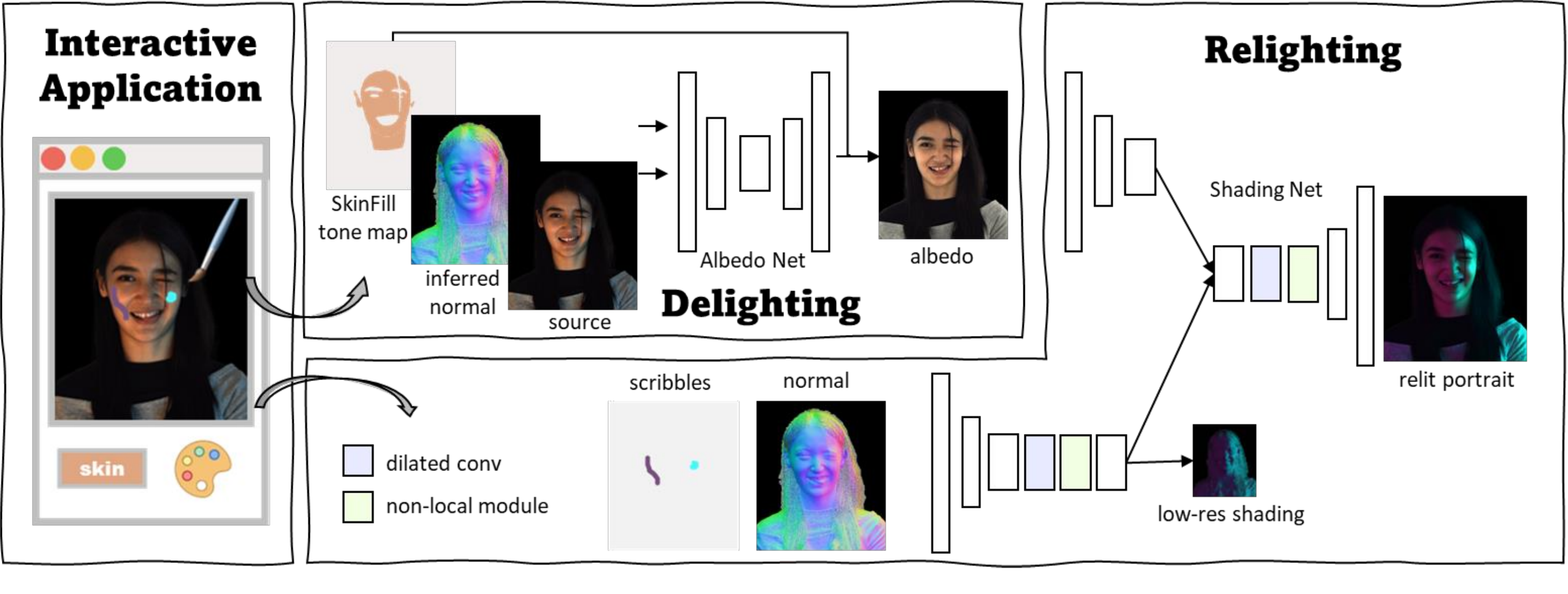}
\vskip-10pt	\caption{\textbf{An overview of LightPainter.} As the user starts scribbling with our interactive application, the neural modules interactively render a realistic relit image that is faithful to the user input. }\label{fig2}
\vspace{-4mm}
\end{figure*}

\paragraph{Portrait Relighting:}
The pioneering work of Debevec \etal~\cite{debevec2000acquiring} presents an advanced illumination rig (i.e. the light stage) to capture per-person reflectance field, which is used to render the subject under novel illuminations. Such technique has been used to create training data for a number of single-image relighting methods~\cite{nestmeyer2020learning,pandey2021total,sun2019single,zhou2019deep,zhang2021neural,zhang2021neural2}. 
Portrait relighting has also been formulated as style transfer. Shih \etal~\cite{shih2014style} employ a multi-scale technique to transfer the local statistics of an exemplar to the target image. Shu \etal~\cite{shu2017portrait} formulate the lighting transfer as a geometry-aware mass transport problem with 3D morphable face model. Using quotient image to achieve relighting is introduced in~\cite{shashua2001quotient,peers2007post}, where they multiply the source image with a ratio map to render novel illuminations. Intrinsic decomposition based approaches~\cite{pandey2021total,wang2020single,Hou_2022_CVPR,nestmeyer2020learning,barron2014shape,le2019illumination,ji2022relight,li2014intrinsic,shahlaei2015realistic} factorize a source image into geometry, reflectance, and illumination, and apply novel lighting by conditioning on a new illumination. Such technique has received increasing popularity over recent years thanks to recently advanced light stage capturing systems~\cite{guo2019relightables} that make direct supervision of intrinsic components possible. Our method also falls into this category.
\vspace{-4mm}
\paragraph{Lighting Representation:} Lighting representation determines how users interact with the system. Several works~\cite{zhou2019deep,sengupta2018sfsnet} use spherical harmonics, which is limited to only low-frequency illuminations. Reference-based methods~\cite{shu2017portrait, shih2014style} use an image as a proxy to represent lighting, yet the requirement of a matching exemplar image reduces their practicability. A similar argument applies to environment map~\cite{sun2019single,zhou2019deep,wang2020single,yeh2022learning,pandey2021total}, which is inherently challenging to edit, thus hard to interactive with. Other works~\cite{nestmeyer2020learning,Hou_2021_CVPR,Hou_2022_CVPR} model only directional lights, which constrains the type of lighting these methods support.  
\vspace{-4mm}
\paragraph{Image Manipulation with User Scribbles.} Scribbling (or sketching) is one of the most intuitive interactions for human to express creative ideas. Drawing-based interface has been widely exploited in various image manipulation tasks~\cite{chen2018sketchygan,nazeri2019edgeconnect,elder1998image,dekel2018sparse,olszewski2020intuitive,yang2020deep,ghosh2019interactive,fivser2016stylit}. For example, the pioneering work of Eitz \etal~\cite {chen2009sketch2photo} introduces Sketch2Photo to interactively perform image retrieval and synthesize from user sketch. Yu \etal~\cite{yu2019free} develops deepfill-v2, which allows users to conduct free-form inpainting with scribbled ``holes''. SketchHairSalon~\cite{xiao2021sketchhairsalon} makes hair design easy by drawing desired hair structures. Scribbles and sketches have also been used for face manipulation~\cite{zeng2022sketchedit,PortenierHSBFZ18} and image colorization~\cite{LevinLW04,HuangTCWW05}. However, such intuitive interface has not been studied in the context of portrait relighting. 
\vspace{-4mm}
\paragraph{Commercial Lighting Editing Tools.}
Only a few commercial applications support a complete set of lighting editing capability. Applications such as Facetune~\cite{Facetune},``Studio Lighting Mode" in iPhone~\cite{apple} and ``Portrait Relighting" feature in Google Pixel~\cite{google} only support a limited set of editing constrained to changing brightness or adding a fixed-color directional light in 2D. The recently released ClipDrop~\cite{ClipDrop} provides more flexible editing by allowing users to place virtual lights with a chosen color, intensity, distance and radius. However, it does not support removing existing illuminations from the scene. Further, to pursue creative lighting effects, it is possible that users have to manually tune multiple lights at the same time. In the user study, we will show this process is difficult for many novices.
\section{Method}
In this section, we describe the framework of LightPainter. As shown in Figure~\ref{fig2}, LightPainter consists of a frontend interactive application for the user to scribble and a backend performing relighting with respect to the user input. A detailed walk-through of the frontend interface can be found in the supplementary. Here we focus our discussion on the backend. Specifically, the backend comprises two conditional neural networks: a skin-tone-conditioned delighting module and a scribble-conditioned relighting module. The delighting stage recovers the geometry representing per-pixel surface normal, and an albedo image that optionally follows a user-selected skin tone. After delighting, estimated normal and albedo are transmitted to the relighting module, which renders the portrait under the lighting condition that respects the user's scribbles. In Section~\ref{sec:delight} and~\ref{sec:relight}, we describe each stage in detail. We define our training objectives in Section~\ref{sec:loss}. 

\begin{figure}[t!]
	\centering
	\includegraphics[clip, trim=0cm 11cm 12cm 0cm, width=1\columnwidth]{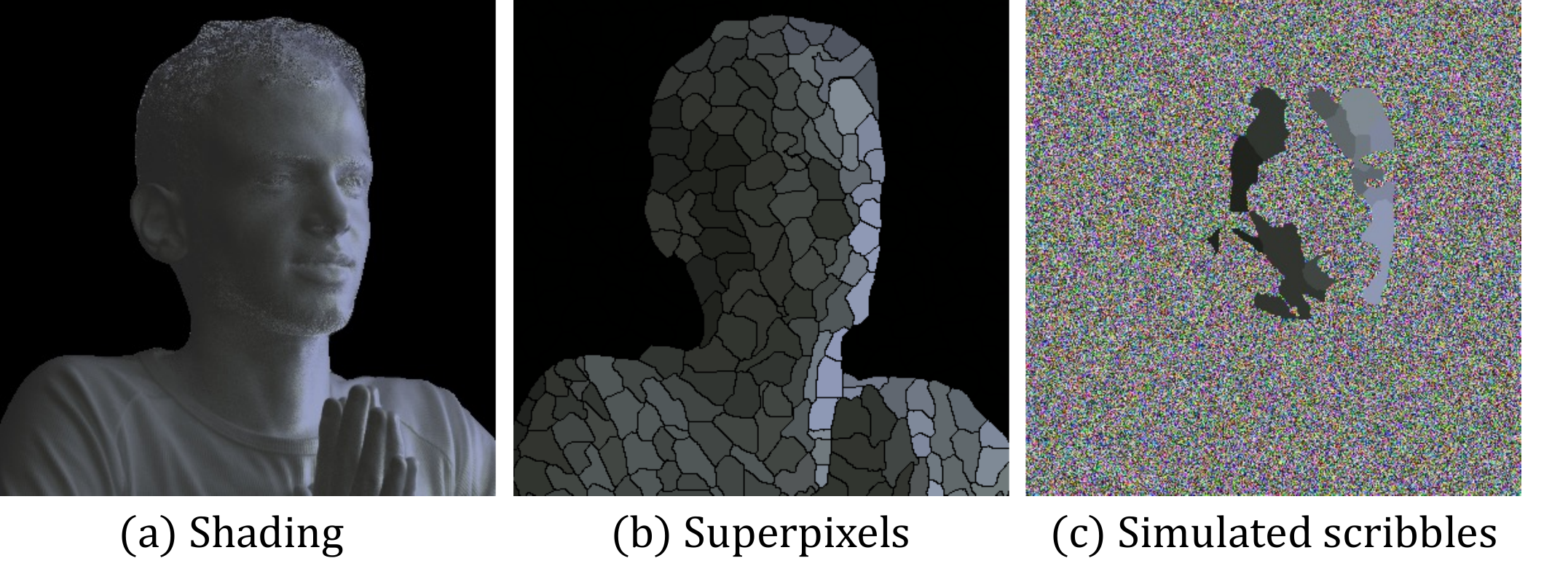}
	\caption{\textbf{An example of the simulated scribbles.} (a) A complete shading obtained from the Phong shading~\cite{phong1975illumination}. (b) Segmented superpixels after quantization and color/intensity average. Each segment is coarse in both shape and intensity. (c) Simulated scribbles generated by sampling from (b). We fill in Gaussian noise into the empty region and background.}\label{simulation}
\end{figure}
\subsection{Delighting Module}\label{sec:delight}
 Inspired by Total Relighting~\cite{pandey2021total}, LightPainter uses two networks to separately estimate geometry and reflectance. For geometry, we use the existing algorithm~\cite{bae2021estimating} and fine-tune on our dataset. Our main difference from prior works is the reflectance prediction, where we propose a skin-tone-conditioned albedo model.

\subsubsection{Skin-tone-conditioned Albedo Net} 
Data-driven albedo prediction is challenging as it requires a fully comprehensive and balanced dataset to avoid any skin tone bias. To address this challenge and recover an accurate albedo, we explore an alternative solution by leveraging user control. This is built upon the observation that the skin tone of a subject can be easily specified in practice, for example, using the skin swatches on a cosmetics website\footnote{For example,
\href{https://www.sephora.com/contentimages/productpages/pdfs2017/StudioUndertoneFinder_v3.pdf}{skin-tone finder} provided by Sephora}. We thus propose to leverage user interaction to aid our albedo generation. Specifically, we ask the user to optionally provide a skin color to the system. As shown in Figure~\ref{fig2}, the network then generates the albedo conditioned on the received color vector {$\hat{v}$}, along with the estimated normal and the source portrait. At training time, $\hat{v}$ can be extracted as the mean skin color from the ground truth albedo. For inference, if {$\hat{v}$} is not provided by the user, the albedo generation scheme falls back to a standard unconditional method.

It is crucial to determine how to best leverage {$\hat{v}$}. Intuitively, the skin tone should be accessible by all pixels in the skin region for guidance. The network must also be designed to follow the guidance so as the generated albedo matches the user's desire. In the following, we introduce a new technique, dubbed \textit{SkinFill}, by drawing inspiration from the makeup routine.
\vspace{-4mm}
\paragraph{SkinFill.} In standard makeup routine, skin color can be modified by blending foundation smoothly over the facial skin, followed by local retouching. This inspires two design choices of SkinFill: (1) matching the exact skin tone by uniformly shifting the pixel value in the skin region, (2) recovering local facial details from the input to ensure fidelity and realism. Specifically, SkinFill first creates a per-pixel skin-tone representation $T$, the \textit{tone map}, by filling in the skin-parsing mask $M_{skin}$ with color $\hat{v}$, i.e. $T = M_{skin}\odot \hat{v}$. The tone map $T$ is then used to condition the network for better facial detail recovery, and directly added to the network prediction to shift the pixel value in the skin region (see the Delighting part in Figure \ref{fig2}). SkinFill brings several benefits:  (1) the tone map makes user guidance easily accessible at all skin pixels. (2) uniformly shifting the skin color towards $\hat{v}$ enforces the prediction to follow the user's intention. (3) The network can focus on recovering the local facial details without the need of regressing the skin tone.

\subsection{Relighting Module}~\label{sec:relight}
In this section, we describe the scribble-based relighting module in detail. To begin with, we introduce a scribble simulation algorithm that enables training a network with synthetic scribbles that resemble real users' inputs. We also describe the shading network that renders a realistic relit portrait conditioned on the input scribbles.
\vspace{-3mm}
\subsubsection{Scribble Simulation} 
In order to train a relighting network that is conditioned on scribbles, we propose a scribble simulation algorithm that automatically generates scribbles that mimic real user inputs and with large variations. 

As aforementioned, we use ``shading stripes" to represent user scribbles, which reflect local shading patterns. Different shading levels naturally correspond to different scribble intensities. Hence the first step in our simulation algorithm is to obtain a full shading map. This is accomplished by rendering with the Phong shading~\cite{phong1975illumination} model using the ground truth geometry and environment map. As shown in Figure ~\ref{simulation} (a), the rendered shading map can be viewed as an ``ideal'' scribble containing detailed and complete lighting information. However, drawings from novices are usually noisy -- irregular and incomplete. To make our system robust to these imperfect inputs, we apply a series of augmentations to model these defects in real scribbles. Specifically, we first convert the shading into $Lab$ color space and ``coarsen'' the luminance channel $L$ by randomly quantizing it into multiple bins. Then we perform a superpixel segmentation using SEEDS~\cite{bergh2012seeds} and average the color within each segment. As shown in Figure \ref{simulation} (b), each segment ends up being coarse in intensity and shape, similar to the noisy scribble that novices tend to draw. Finally, we ``sparsify'' the simulated scribbles by randomly sampling a small subset of segments at each training step. The sampling rate is drawn from a truncated exponential distribution with $\lambda=3$, which results in mostly sparse inputs. In addition, we always keep the segments of top $5\%$ brightest and darkest intensity to make sure our sampling captures the full dynamic range of shading. On the other hand, this also ensures that the most representative lighting information is preserved to help the network reasonably complete the full shading map. An example of simulated scribbles is illustrated in Figure~\ref{simulation} (c). More details and hyper-parameter choices can be found in the supplement. 

\subsubsection{Scribble-conditioned Shading Net}
\paragraph{Two-Step Relighting Pipeline.} In prior works, relighting is often performed in a single step by taking the lighting condition as a conditional input. In our interactive setting, the lighting condition is from the user scribbles, which can be local, sparse and coarse. We found that directly predicting the relit images from the scribbles does not perform well, where the network would struggle at generating a plausible completion of shading.

To address this challenge, we introduce a two-step relighting pipeline: The first step completes shading following the subject geometry, and the second stage refines the completed shading to render a relit image. As shown in Figure~\ref{fig2}, we use the bottom branch to complete a low-resolution shading map conditioned on the normal and scribbles. The output is supervised with the ground truth shading, which enforces the network to learn to propagate sparse lighting information following the surface geometry, which encourages geometry-consistent relighting. The shading feature is then concatenated with the albedo feature (encoded by the upper brunch), and transmitted to the decoder. The decoder then refines the completed shading and renders the final image.
\vspace{-3mm}
\paragraph{Improved Network Architecture.} We adapt our network architecture to better tolerate ``incomplete" user scribbles. Intuitively, the receptive field of the network should be sufficiently large so that the network can leverage the global context given sparse and local user input. To achieve this, we build our network with a U-Shaped structure~\cite{ronneberger2015u} and further adapt it with additional dilated convolutions~\cite{yu2016dilated} and non-local modules~\cite{wang2018non}. These improvements allow a global receptive field and enhance the information flow among distant locations, thus are more suitable for our task. We will demonstrate that our architectural design is crucial for faithful relighting from sparse scribbles.
\subsection{Training Objective}~\label{sec:loss}
To ensure both realism and fidelity of the relit image,
our neural network optimizes the following training objectives: 

\textbf{Reconstruction loss} $\m{L}_{R_{alb}}$ and $\m{L}_{R_{relit}}$: the standard $L1$ distance between the generated albedo/portrait and the ground truth albedo/portrait to ensure content fidelity. 

\textbf{Perceptual loss~\cite{zhang2018unreasonable}} $\m{L}_{P_{alb}}$ and $\m{L}_{P_{relit}}$: the features-wise distance of the predicted albedo/portrait and ground truth albedo/portrait extracted by a pre-trained VGG~\cite{vgg}. This loss is used to improve visual quality.

\textbf{Shading reconstruction loss} $\m{L}_{R_{shad}}$: the standard $L1$ distance between the completed low-resolution shading and the downscaled ground truth to enforce shading completion.

The overall loss can be expressed
as follows:
\begin{align}
\vspace{-2mm}
    \m{L} = \m{L}_{R_{alb}}+\m{L}_{P_{alb}}+\m{L}_{R_{relit}}+\m{L}_{P_{relit}}+\m{L}_{R_{shad}}
\end{align}

\section{Data and Implementation Details}
\paragraph{Data Preparation.} Following prior practice ~\cite{pandey2021total,sun2019single,zhang2021neural}, we use a light stage~\cite{debevec2000acquiring} to collect training and testing OLAT data. Our capture system is structurally similar to that used in~\cite{sun2019single}. Our light stage contains 160 programmable LED-based lights and 4 frontal-viewing high-speed cameras. The detailed configuration can be found in the supplementary material. Our dataset contains 59 subjects. Each subject is photographed with 5-15 different poses and accessories, resulting in 2123 OLAT sequences in total. A subset of 13 subjects with diverse races and different genders are used for testing. The ground truth normal is computed following the algorithm described in~\cite{pandey2021total}. The ground truth diffuse albedo is acquired by capturing the subject in a flat unidirectional lighting condition with all LED lights turned on.

To obtain a paired dataset for supervised training, we render our OLAT images under diverse lighting environments, which are collected from the Laval Indoor \cite{gardner2017learning} and Outdoor HDR datasets and PolyHaven \cite{ployhaven}. We collect in total 2571 real environment maps. We in addition create 2000 synthetic environment maps by placing colored eclipse shapes on a black canvas. We randomly select a subset of 450 environment maps for testing. We also augment lighting by randomly rotating the environment maps when rendering data, resulting in 870K training samples. For the test set, we randomly pair each test OLAT sequence with two lighting environments to form input and output pairs, resulting in 757 testing pairs.
\begin{figure*}[t!]
	\centering
	\includegraphics[clip, trim=0cm 2.5cm 11cm 0cm, width=0.97\textwidth]{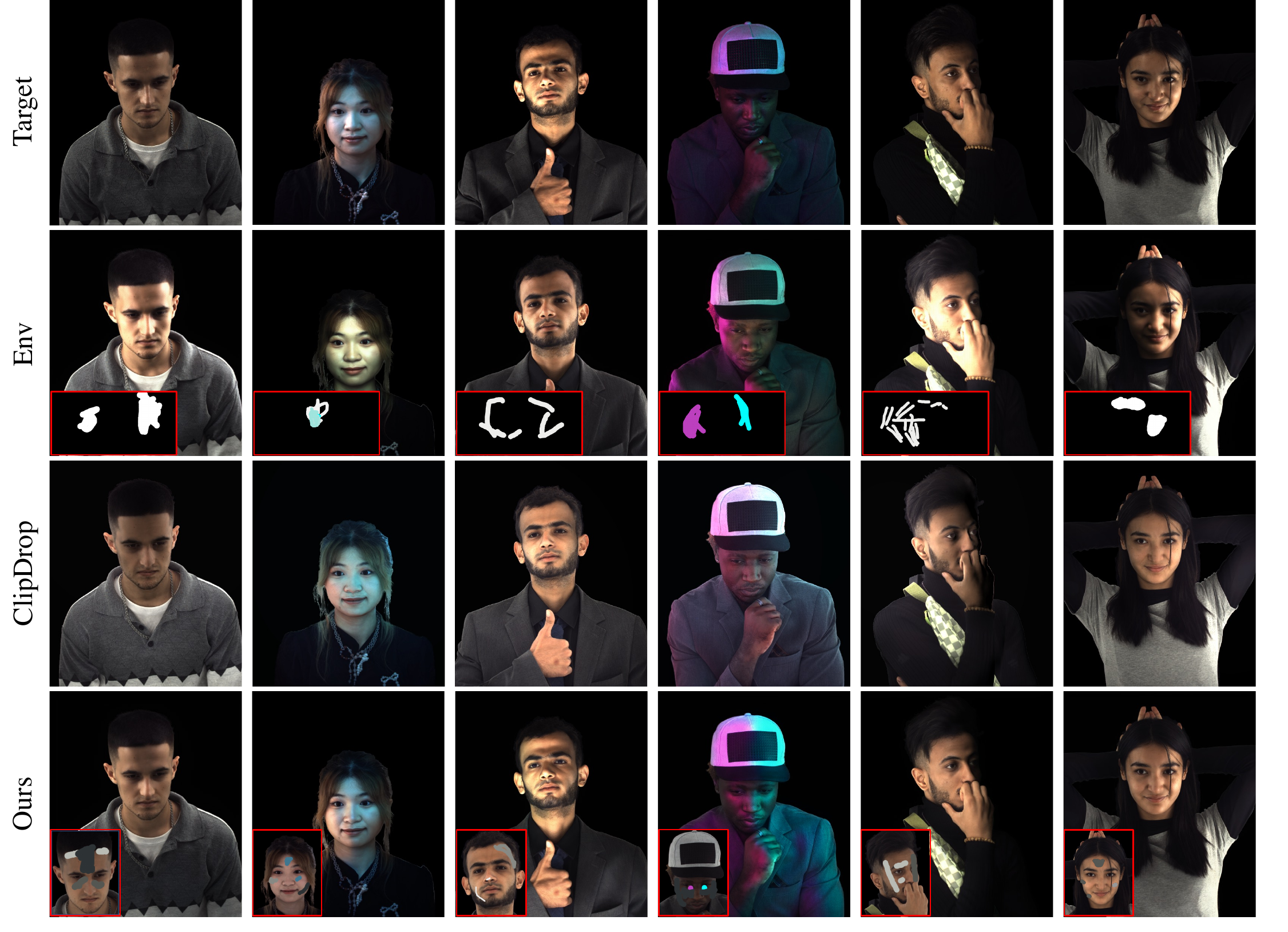}
\vskip-8pt	\caption{\textbf{Visual comparisons on user-generated images using three relighting systems.} User-drawn scribbles and environment maps are shown as insets. Lighting effects that are produced by our method are the most faithful and consistent with the target. \textbf{Best viewed by zooming to 4X.}}\label{fig:user}
\vspace{-5mm}
\end{figure*}
\vspace{-4mm}
\paragraph{Implementation and Training Details.} Both Shading Net and Albedo Net have an encoder-decoder structure with three downsampling and upsampling layers. Multiple standard convolutions, dilated convolutions~\cite{yu2016dilated} and non-local attentions~\cite{wang2018non} are inserted at the bottleneck of each network. Further details can be found in the supplementary material. For training, we resize the rendered data to $800\times 600$ resolution and randomly crop $512\times512$ region from 32 images to form a mini-batch. The network is trained using Adam optimizer~\cite{adam}. The learning rate is set to 1e-4 for the first 2 epochs, then reduced to a half after each epoch. We stop the training after 5 epochs. The proposed model is implemented using PyTorch and the training takes about 1 day on 8 Nvidia A100 GPUs.
\section{Experiments}
\vspace{-1mm}
We now demonstrate the high-quality portrait relighting capability of LightPainter via extensive evaluations, comparisons, and user studies. We also provide ablation studies to demonstrate the benefits of our system design.

\textbf{Evaluation Metrics.} We report perceptual metrics LPIPS~\cite{zhang2018unreasonable}, NIQE~\cite{zhang2015feature}, and pixel similarity metrics PSNR and SSIM~\cite{wang2004image}. In addition, we use the Deg (cosine similarity between LightCNN~\cite{wu2018light} features) to evaluate identity preservation capability. All results are computed within the subject mask pre-computed using~ \cite{yu2021mask}.

\subsection{User Study}
To demonstrate LightPainter can
benefit general users on portrait lighting editing, we perform a user study to evaluate the quality and user experience of LightPainter.
We recruit general users to conduct portrait relighting with three systems that use different interactive approaches:
\begin{itemize}[noitemsep]
\vspace{-1.5mm}
    \item ClipDrop~\cite{ClipDrop}\footnote{www.clipdrop.co/relight}, a commercial lighting editing web service, where users can place virtual lights into the scene with a chosen light color, intensity, distance and radius.
    \item Env: a re-implementation of Total Relighting~\cite{pandey2021total}, the state-of-the-art portrait relighting method. With this tool, the user provides a hand-drawn ``environment map'' to perform lighting editing.
    \item LightPainter, our scribble-based system.
\vspace{-1mm}
\end{itemize}

\begin{figure}[t]
\begin{center}
\includegraphics[clip, trim=0cm 10cm 19.5cm 0cm, width=0.48\textwidth]{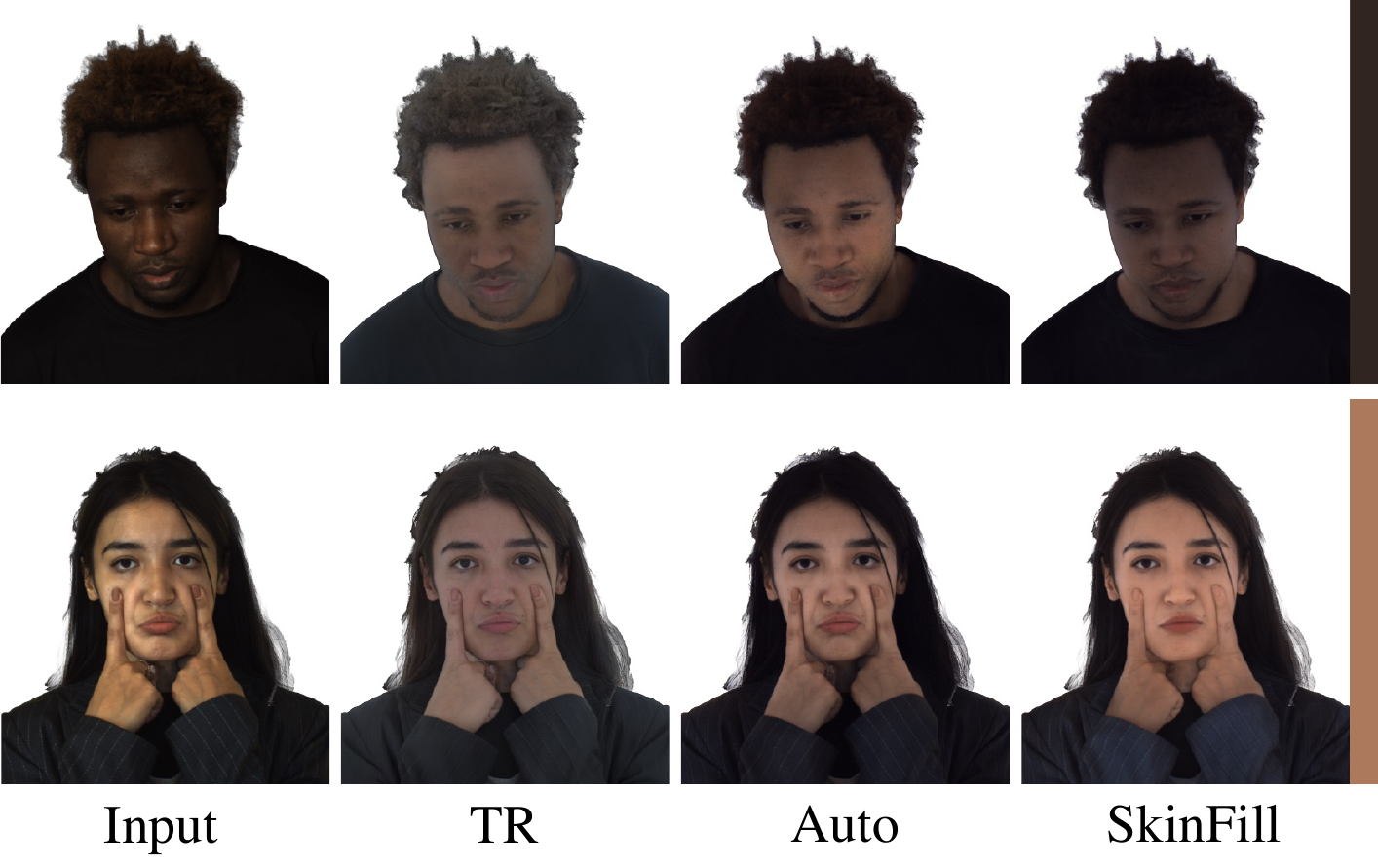}
\end{center}
\vspace{-14pt}
\caption{\textbf{Examples of how user leverage SkinFill on skin color retouching.} ``Auto'' denotes albedo generated by the automatic prediction mode of the proposed LightPainter. The selected skin color swatch is appended at the end of each row. Both TR \cite{pandey2021total} and ``Auto'' suffer from skin-tone bias. In contrast, with SkinFill, LightPainter can predict the correct skin color that follows the user's desire.}
\label{fig:skin}
\vspace{-5mm}
\end{figure}

\begin{table}[t]\label{tab:user}
       
        \small
        \centering
        \caption{Quantitative evaluation on user-generated images.}
         \vspace{-2mm}
        \tabcolsep=0.1cm
        \resizebox{\columnwidth}{!}{
                \begin{tabular}{l|cc|c|cc}
                        \hline
                        Methods       & LPIPS$\downarrow$   &NIQE$\downarrow$    & Deg$\uparrow$   & PSNR$\uparrow$  & SSIM$\uparrow$  \\ \hline \hline
                        ENV          &0.1359 &6.067 &0.8939 &20.06  &0.7832 \\ 
                        ClipDrop~\cite{ClipDrop}            &0.1435   &6.967 &0.8917 & 20.49 &0.6116  \\ 
                        \textbf{LightPainter (ours)} & \textbf{0.0868}  & \textbf{5.643}& \textbf{0.9379}&\textbf{24.94}  &\textbf{0.8373} \\ \hline
                        
        \end{tabular}}\label{tab:user}
        \vspace{-3mm}
\end{table}


For this experiment, we provide a \textit{target image} with a random lighting effect and assign a task to the users to reproduce this lighting effect on an input \textit{source image} with each tool individually. The source image and target image are from our testing set and only differ in lighting. We impose a time limit of 2 minutes per task for each tool.
Note that ClipDrop does not remove existing lighting effects on source images, we use the albedo image (from our test set) as the source image for fair comparisons. 


In Figure~\ref{fig:user}, we show a random subset of visual results from the user study. Using Env or ClipDrop~\cite{ClipDrop}, users are often unable to faithfully reproduce the target lighting effects. We notice that results from using Env exhibit a significant mismatch in both brightness and shading patterns. This shows the difficulty for users to interpret the lighting position and intensity, and associate lighting effects on an image with environment map representations. While ClipDrop~\cite{ClipDrop} is easier for users to interact with, the desired detailed lighting patterns in the target image, such as the highlight, are still largely absent in the results. 
In contrast, LightPainter allows users to faithfully reproduced the target lighting effect, with convincing details, within two minutes.

We also provide quantitative evaluations (computed between user-generated results and target images) on each method, as shown in Table~\ref{tab:user}. The quantitative results are collected on 40 trials performed by 20 users.
LightPainter achieves the best performance on all metrics, which not only produces the target lighting most closely, but also shows the best image quality and identity preservation capability.

\begin{figure}[t]
\begin{center}
\includegraphics[clip, trim=0cm 1.5cm 19.5cm 0.3cm, width=0.49\textwidth]{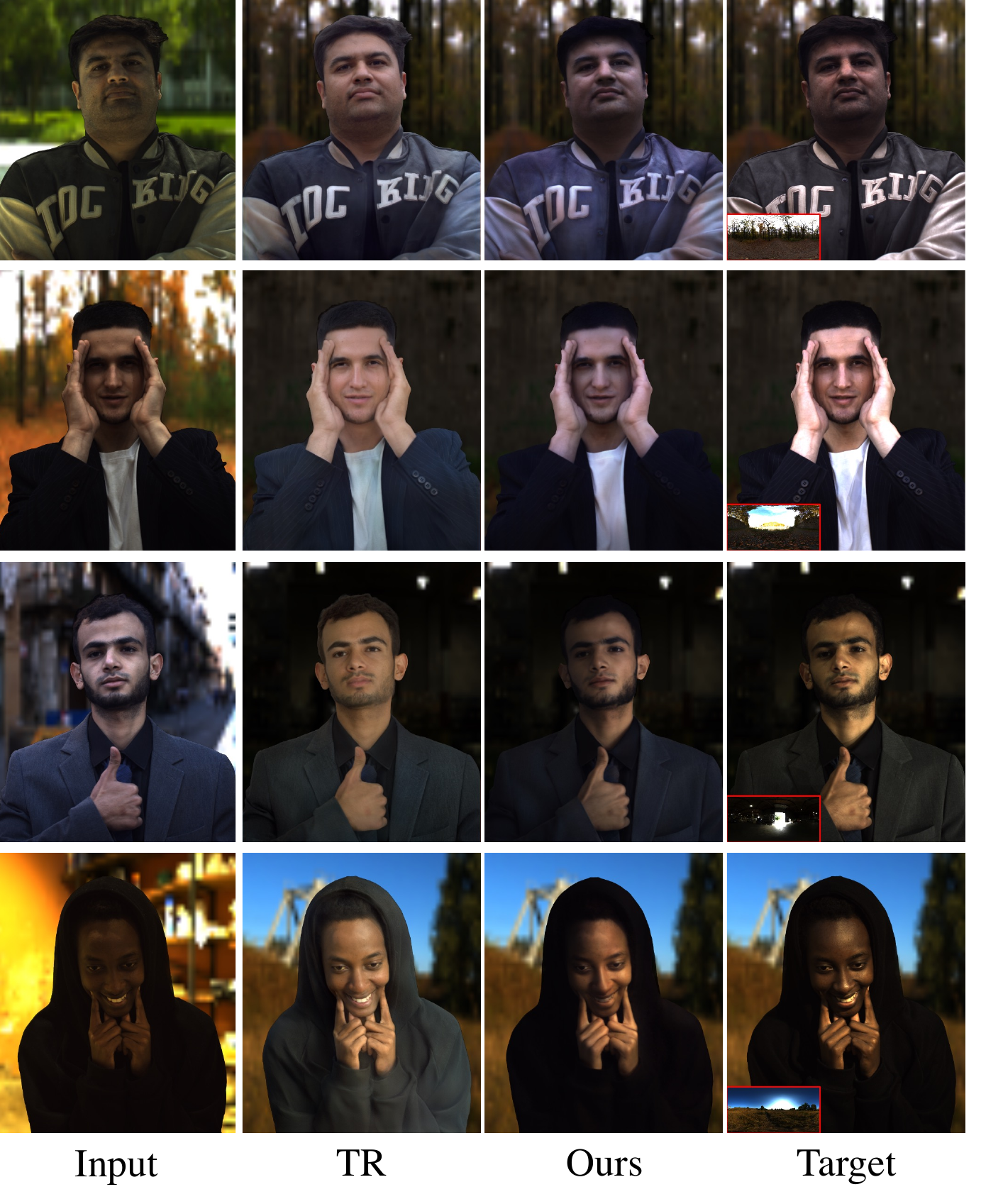}
\end{center}
\vspace{-7mm}
\caption{\textbf{Qualitative comparisons on environment-map-based portrait relighting.} Our method can produce on-par or better results comparing with TR~\cite{pandey2021total}. Inputs and targets (ground truth) are generated using the test set of light stage subjects and environment maps not seen during training. 
}
\vspace{-5mm}
\label{fig: pr}
\end{figure}

We gathered feedback from the 20 participated users on their experience with each relighting system. We summarize the findings as follows: 1. All users are satisfied with our scribble-based relighting scheme, and feel that drawing over the portrait is very easy to operate. 2. Most users (17/20) feel that editing/drawing environment map is confusing. 3. More than half of users (12/20) feel tuning virtual lights (i.e. guessing the color, intensity and distance of each light associated with the scene) is very difficult and it is hard to obtain the desirable lighting effect. 

\begin{figure}[h]
\begin{center}
\includegraphics[clip, trim=0.1cm 23.5cm 16cm 0.0cm, width=0.49\textwidth]{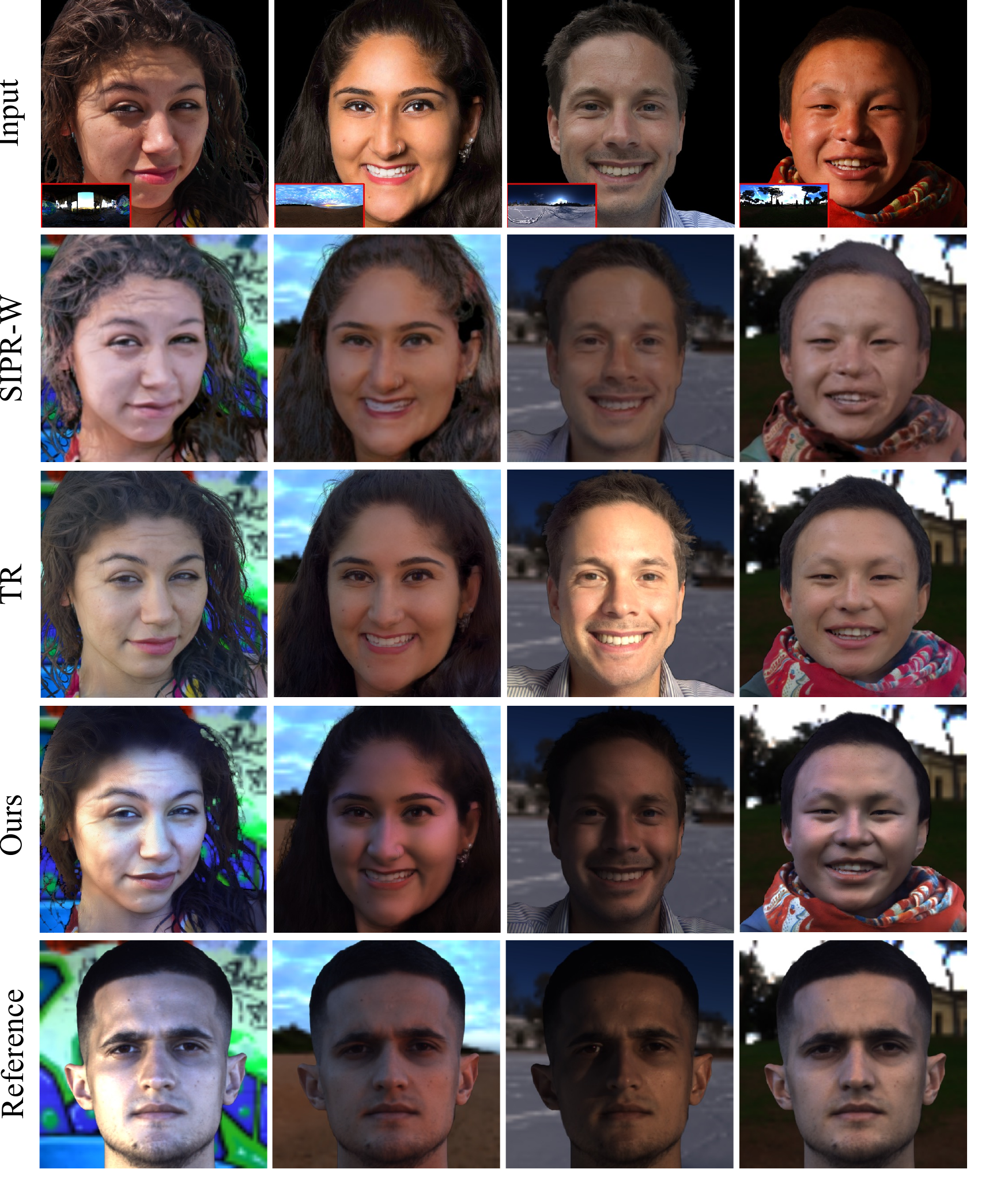}

\end{center}
\vspace{-6mm}
\caption{\textbf{Qualitative comparisons on in-the-wild face relighting.} We compare relighting results with SIPR-W \cite{wang2020single} (row-2) and TR~\cite{pandey2021total} (row-3). The environment maps are shown as insets (row-1). We provide a reference image (row-5) rendered with OLAT data as guidance of the lighting effect under the input environment.
}
\vspace{-6mm}
\label{fig: ffhq}
\end{figure}

\subsection{Skin-tone Control}
With SkinFill, LightPainter can predict the albedo that respects the user-specified skin color. This grants the user with the flexibility of tuning skin tone.
We demonstrate its use case in Figure~\ref{fig:skin}. This also helps resolve the potential data bias and ambiguity in predicting an accurate skin tone from a single image. As already shown in machine learning based approaches, the skin color of the predicted albedo from both total relighting~\cite{pandey2021total} and the automatic mode of LightPainter is inaccurate and appears washed-out. In contrast, with additional user interference with SkinFill, LightPainter can produce a more faithful albedo that respects the user's intent. More experiments can be found in the supplement.

\subsection{Comparisons with State-of-the-art Methods}
To further demonstrate the relighting quality of LightPainter, we compare it with the state-of-the-art methods SIPR-W~\cite{wang2020single} and Total Relighting~\cite{pandey2021total} (TR) on environment-map-based relighting. Since LightPainter is not designed for using environment map as its lighting representation, we perform relighting with ``estimated" scribbles instead of user scribbles.
Specifically, we render the shading map using the Phong model~\cite{phong1975illumination} with the estimated normal map to replace the user scribbles. 
These estimated scribbles are in fact ``ideal'' inputs for LightPainter, with more complete shading information than either the simulated scribbles we used for training or the real user scribbles.
We adopt the off-the-shelf image matting method~\cite{yu2021mask} to extract the foreground portrait and composite it with the new background.
We report (1) quantitative and qualitative results on portrait/upper-body relighting on our testing data, and (2) qualitative comparisons on in-the-wild images from FFHQ~\cite{karras2019style}. 
Results of SIPR-W~\cite{wang2020single} and TR~\cite{pandey2021total} are directly obtained from their authors.

\vspace{-4mm}
\paragraph{Evaluation with Light Stage Data.} For this experiment, 757 ground truth images are rendered with environment maps and OLAT data from the test set using Image-base Relighting~\cite{debevec2000acquiring}. We report the quantitative results in Table~\ref{tab:compare} and provide comparisons. Our approach performs better in perceptual quality, identity preservation, and image similarity.
In Figure~\ref{fig: pr}, we show qualitative comparison with~\cite{pandey2021total}. Our method produces photo-realistic results and respects the target lighting more faithfully.
\vspace{-4mm}
\paragraph{Evaluation with In-the-wild Images.} 
We demonstrate in-the-wild portrait relighting capability with images from FFHQ~\cite{shih2014style} dataset.
We show qualitative results in Figure~\ref{fig: ffhq}. Since there is no relighting ground truth, we provide a reference image by rendering a face under the target lighting environment using OLAT and Image-base Relighting~\cite{debevec2000acquiring}.
LightPainter generates high-quality relighting results with convincing lighting effects. Our results are more consistent with lighting effects in the reference images compared to TR~\cite{pandey2021total}. Compare to~\cite{wang2020single}, our results are more robust and exhibit no noticeable artifacts.

\begin{table}[t]
       
        \small
        \centering
        \caption{Quantitative comparsion with Total Relighting (TR).}
         \vspace{-3mm}
        \tabcolsep=0.1cm
        \resizebox{\columnwidth}{!}{
                \begin{tabular}{l|cc|c|cc}
                        \hline 
                        Methods       & LPIPS$\downarrow$   &NIQE$\downarrow$    & Deg.$\uparrow$   & PSNR$\uparrow$  & SSIM$\uparrow$  \\ \hline \hline
                        TR~\cite{pandey2021total}       &0.2160 &8.137 &0.8454 &20.31 &0.5705 \\ 
                        
                        \textbf{LightPainter} & \textbf{0.1383}  & \textbf{6.195}& \textbf{0.9076}&\textbf{25.71}  &\textbf{0.8449} \\ \hline
                        
        \end{tabular}}
        \label{tab:compare}
        \vspace{-3mm}
\end{table}

\begin{table}[t]
       
        \small
        \centering
        \caption{Ablation study on enlarged receptive field.}
         \vspace{-3mm}
        \tabcolsep=0.1cm
        \resizebox{\columnwidth}{!}{
                \begin{tabular}{l|cc|c|cc}
                        \hline
                        Methods       & LPIPS$\downarrow$   &NIQE$\downarrow$    & Deg.$\uparrow$   & PSNR$\uparrow$  & SSIM$\uparrow$  \\ \hline \hline
                        
                        One-Step          &0.1423 &7.691 &0.8874 &23.88 &0.5278 \\ \hline 
                        w/o Non-Local          &0.0875 &7.266 &0.9170 &27.46  &0.8152 \\ 
                        w/o dilated conv          &0.1020   &8.318 &0.9169 &27.18 &0.8777  \\ \hline 
                        \textbf{LightPainter} & \textbf{0.0848}  & \textbf{7.012}& \textbf{0.9310}&\textbf{28.48}  &\textbf{0.8899} \\ \hline
                        
        \end{tabular}
        }
        \label{tab:arch}
        \vspace{-5mm}
\end{table}

\subsection{Ablation Study}
We now provide ablation studies to demonstrate the benefit of our key designs in LightPainter. All results are evaluated on our test set using synthetic scribbles, ground-truth albedo and normal with a fixed sampling ratio of $0.3$.
\vspace{-4mm}
\paragraph{Effectiveness of Two-step Relighting Scheme.}
Our system adopts a two-step relighting scheme to enforce geometry-consistent shading. We investigate its effectiveness by comparing it with a single-step baseline, which directly predicts the relit images from the scribbles without shading completion. The results for one-step baseline are reported in the first row of Table \ref{tab:arch}. As shown, two-step approach (i.e. LightPainter) significantly improves performance.
\vspace{-4mm}
\paragraph{Augmented Receptive Field.} 
To handle sparse user input, we adopt non-local blocks~\cite{wang2018non} and dilated convolutions~\cite{yu2016dilated} to enlarge the receptive field of the Shading Net. We conduct experiments to validate this design choice. As shown in Table~\ref{tab:arch}, removing either non-local attention or dilate convolution harms performance.

\section{Conclusion}


In this paper, we introduce LightPainter, a novel interactive and intuitive portrait relighting system that uses free-hand scribbles as the user interface. To address the challenges of relighting with scribbles, we propose novel network designs and a shading-based scribble simulation approach to generate training data.
We also introduce a conditional delighting module that predicts high-quality albedo optionally conditioned on skin tone. Experiments and user study demonstrate the state-of-the-art portrait relighting performance of our method. Noticeably, LightPainter allows users to create desirable portrait lighting effects with ease. Discussion of limitations can be found in the supplementary material.



\vspace{-5mm}
\paragraph{Acknowledgments}
This work was supported by NSF
CARRER award 2045489. We thank Chaowei Company for the support of light stage data.

{\small
\bibliographystyle{ieee_fullname}
\bibliography{egbib}
}

\end{document}



\title{Supplementary File:
\\ LightPainter: Interactive Portrait Relighting with Freehand Scribble}

\author{%
  Yiqun~Mei$^{1}$\quad He Zhang$^{2}$\quad Xuaner Zhang$^{2}$\quad Jianming Zhang$^{2}$\quad Zhixin Shu$^{2}$\quad Yilin Wang$^{2}$\quad\\ Zijun Wei$^{2}$\quad Shi Yan$^{2}$\quad HyunJoon Jung$^{2}$\quad Vishal M.~Patel$^{1}$ \\ \\
  {\small$^{1}$Johns Hopkins University\quad \quad $^{2}$Adobe Inc.}
}

\maketitle
\begin{figure}[t]
    \centering
    \includegraphics [width=0.47\textwidth]{figure_supp/lightstage.jpg}
    \vspace{-3mm}
    \caption{An illustration of our light stage system.}
    \vspace{-3mm}
    \label{fig:lightstage}
\end{figure}

\begin{figure}[h]
    \centering
    \includegraphics[width=0.47\textwidth]{figure_supp/lightstage.png}
    \vspace{-1mm}
    \caption{The LED light distribution of our light stage system. The image is shown in the panoramic format with $x$-axis denoting the longitude and $y$-axis denoting the latitude.}
    \vspace{-5mm}
    \label{fig:lightstage_light}
\end{figure}

\begin{figure}[t]
    \centering
    \includegraphics[clip, trim=0 13.9cm 15.8cm 0, width=0.5\textwidth]{figure_supp/olat_exp.pdf}
     \vspace{-8mm}
    \caption{Examples of the captured OLAT images from 4 frontal-view cameras}
     \vspace{-5mm}
    \label{fig:olat_exp}
\end{figure}

\begin{figure}[ht]
    \centering
    \includegraphics[clip, trim=0 11cm 7.8cm 0, width=0.48\textwidth]{figure_supp/skin_tone_adjust.pdf}
    \vspace{-7mm}
    \caption{Examples of skin-tone control. Given a source image (left), LightPainter can predict the albedo that respects user-specified skin color (bottom left).}
    \vspace{-3mm}
    \label{fig:skin}
\end{figure}
\begin{figure}
    \centering
    \includegraphics[clip, trim=0 8.9cm 10cm 0cm, width=0.49\textwidth]{figure_supp/limitations.pdf}
    \vspace{-6mm}
    \caption{Limitations of our method. While novice users can easily transfer the lighting patterns from a target image by simple scribbling, the scribble-based interface does not allow professional users to specify the exact shape/boundary of shadows. In LightPainter, these details are handled by the network to tolerate imperfect scribbles from novices. In addition, as shown in the relit image, our network also failed to produce non-Lambertian reflections in the subject's eyes.}
    \label{fig:limitation}
\end{figure}
\section{Interactive Application}
 The interactive application of LightPainter implements drawing tools for users to draw and edit their scribbles with ease. In the supplement, we provide a ~\textbf{recorded video} of using LightPainter to perform in-the-wild portrait relighting to demonstrate our interface and workflow.

\section{Implementation Details}

\begin{figure*}[ht]
    \centering
    \includegraphics[clip, trim=0 40.5cm 26cm 0, width=\textwidth]{figure_supp/supp_figure_wild.pdf}
    \vspace{-10mm}
    \caption{Examples of relighting in-the-wild portraits from user scribbles}
    \vspace{-2mm}
    \label{fig:portrait_visual_draw}
    \end{figure*}

\begin{figure*}[ht]
    \centering
    \includegraphics[clip, trim=0 5.5cm 6cm 0, width=\textwidth]{figure_supp/supp_more_visual_results_for_re.pdf}
    \vspace{-10mm}
    \caption{More visual results on images with eyeglasses, light hair color and three color lighting.}
    \vspace{-2mm}
    \label{fig:portrait_visual_draw_more}
    \end{figure*}
    
\paragraph{Unconditional Albedo Net.} We also train an unconditional
albedo net using the same architecture, which takes only
the portrait image and its normal as input. If the skin color
is not provided, our backend will automatically switch to
this unconditional network for inference.

\paragraph{Network Architecture.}
In LightPainter, both Albedo Net and Shading Net adopt a U-Shaped structure. Specifically, Albedo Net is a standard UNet~\cite{ronneberger2015u} and Shading Net has one additional non-local layer~\cite{wang2018non} and dilated convolution~\cite{yu2016dilated} in the bottleneck of both the encoder (shading completion branch) and the decoder. For both networks, we use 3 upsampling layers in the encoder and 3 downsampling layers in the decoder, resulting in 64-128-256-512 and 512-256-128-64 hidden channels respectively. All latent features in the bottleneck have 512 channels.  The upsampling operation is implemented using transposed convolution with a kernel size of $6\times6$. The dilated convolutions have a dilation rate of $7\times7$. And the kernel size for standard convolutions is set to $3\times3$.

\begin{figure*}[ht]
    \centering
    \includegraphics[clip, trim=0 2cm 15.5cm 0, width=\textwidth]{figure_supp/main_fig_supp_2.pdf}
    \vspace{-7mm}
    \caption{Visual comparisons on environment-map-based portrait relighting. We compare our method with Total Relighting~\cite{pandey2021total} (TR) and SIPR-W~\cite{wang2020single}.}
    \label{fig:portrait_visual_1}
\end{figure*}

\begin{figure*}[ht]
    \centering
    \includegraphics[clip, trim=0 2cm 15.5cm 0, width=\textwidth]{figure_supp/main_fig_supp_1.pdf}
    \vspace{-7mm}
    \caption{Visual comparisons on environment-map-based portrait relighting. We compare our method with Total Relighting~\cite{pandey2021total} (TR) and SIPR-W~\cite{wang2020single}.}
    \label{fig:portrait_visual_2}
\end{figure*}

\paragraph{More Scribble Simulation Details.} To compute shading maps, we follow Total Relighting~\cite{pandey2021total} and use the prefiltering technique for environment-map-based Phong shading. We pre-compute the diffuse and specular irradiance maps for each environment map by filtering it with cosine lobe functions corresponding to Lambertian and
Phong specular BRDF (where the Phong exponent is set to 32). This bakes the \textit{ambient} light term into the diffuse
term. The \textit{diffuse} and \textit{specular} shading of the subject are obtained by looking up into these irradiance maps according to the normal or reflection directions, respectively. The final shading
is: $0.85 \times \textit{diffuse} + 0.15 \times \textit{specular}$.

We provide more details about the quantization process in our scribble simulation algorithm. As discussed, we first convert the obtained shading into $Lab$ space and quantize the luminance into multiple bins. We empirically set the number of bins to 25 to ensure the simulated scribbles are sufficiently coarse in terms of intensity so that can mimic causal user input. However, quantization inherently limits the network to be trained on only those 25 predefined bin levels and may result in poor generalization for other scribble intensities. To address this, we propose to shift the bin levels by a random value $p$ at each training step, where $p$ is drawn from $0$ to the bin width. With the random shift, the simulated scribbles can cover the full range of input intensity.

\paragraph{Light Stage Configurations.}
As discussed, we rely on the light stage~\cite{debevec2000acquiring} to prepare our training data. Specifically, we construct a light stage illumination system with a diameter of 3.6m and 160 LED lights. The system structure and the distribution of lights are illustrated in Figure ~\ref{fig:lightstage} \& \ref{fig:lightstage_light}. We use the MER2-502-79U3M high-speed camera to record the reflectance field of the subject at 5 megapixel resolution and exposure time of 20ms. We provide some examples of the captured OLAT images from 4 frontal camera views in Figure \ref{fig:olat_exp}.
\section{More Study on Skin-tone Control}
As discussed, LightPainter can generate an albedo image that respects the user-specified skin color. We demonstrate its effectiveness in terms of user control in Figure ~\ref{fig:skin}. As one can see, our approach can predict realistic albedo images that match the user's intent.

\section{Limitations}
LightPainter is not without limitations. First, similar to previous relighting methods~\cite{pandey2021total,sun2019single,zhang2021neural, zhang2021neural2}, our method is trained on the data rendered with image-based relighting~\cite{debevec2000acquiring}. This inherently limits the expressiveness of the model to the lighting patterns that can be represented by the environment maps, indicating it cannot handle lighting effects and cast shadows caused by occlusions. Second, similar to Total Relighting~\cite{pandey2021total}, it fails to generate specular reflections in the eyes of the subjects (see Figure~\ref{fig:limitation}). This is because the eye regions only contribute to a small portion of the overall loss function. One possible solution is to add explicit supervision on the eye regions and we leave this extension for future work.  Third, in a few examples, we found that some fine details of the hair region are smoothed out. We suspect it may be due to simply fine-tuning the off-the-shelf normal net is not sufficient to estimate the fine and dense geometry of hair strands, and it is also difficult to learn sophisticated light
transport in hairs without explicit supervision. Finally, as discussed, by design our method is robust to noisy and incomplete scribbles from novices, where the plausible completion of the lighting effects, such as sharpness of highlights and hardness of shadow boundary, are left to the network. Therefore, as shown in Figure~\ref{fig:limitation}, although our method is good at producing realistic light patterns that match the user's desires, the scribble-based interface does not allow professional users to specify the lighting details, for example, the exact shape and boundary of shadows. We believe extending the current scribble-based interface to support more fine-grained control for professional users will be an interesting direction for future exploration.

\section{More Visual Results}
In Figure ~\ref{fig:portrait_visual_draw} \&~\ref{fig:portrait_visual_draw_more}, we provide more examples of in-the-wild portrait relighting to better demonstrate the effectiveness of LightPainter. In Figure \ref{fig:portrait_visual_1} \& \ref{fig:portrait_visual_2}, we also provided additional qualitative comparisons with state-of-the-art methods~\cite{wang2020single,pandey2021total} on environment-map-based portrait relighting. Note that we did not compare SIPR-W~\cite{wang2020single} on this task in the main paper as it yields relit results with strong artifacts (because SIPR-W~\cite{wang2020single} is only trained for face images). These results are also included here.

{\small
\bibliographystyle{ieee_fullname}
\bibliography{egbib}
}